\documentclass[journal]{IEEEtran}

\usepackage[T1]{fontenc}
\usepackage{graphicx}
\usepackage{amsmath,amssymb,mathtools}
\usepackage{booktabs}
\usepackage{array}
\usepackage{multirow}
\usepackage[table]{xcolor}
\usepackage{float}
\usepackage{url}
\usepackage{listings}
\usepackage[numbers,sort&compress]{natbib}
\usepackage[colorlinks=true,urlcolor=blue,linkcolor=blue,citecolor=blue]{hyperref}
\lstdefinestyle{schema}{basicstyle=\ttfamily\footnotesize,columns=fullflexible,
  keepspaces=true,breaklines=true,breakatwhitespace=true,frame=tb,
  framerule=0.4pt,xleftmargin=0pt,aboveskip=4pt,belowskip=4pt,
  comment=[l]{//},commentstyle=\color{gray}\itshape,showstringspaces=false}
\allowdisplaybreaks
\newcommand{\bibdoi}[1]{\href{https://doi.org/#1}{\urlstyle{same}\nolinkurl{#1}}}
                          \usepackage{eso-pic}
\AddToShipoutPictureFG*{\AtPageLowerLeft{\put(\LenToUnit{\dimexpr(\paperwidth-\textwidth)/2\relax},\LenToUnit{0.25in}){\parbox[b]{\textwidth}{\centering\footnotesize This work has been submitted to the IEEE for possible publication.\\ Copyright may be transferred without notice, after which this version may no longer be accessible.}}}}
                 \newsavebox{\fittabbox}                        \newcommand{\fittab}[1]{\sbox{\fittabbox}{#1}\ifdim\wd\fittabbox>\linewidth\resizebox{\linewidth}{!}{\usebox{\fittabbox}}\else\usebox{\fittabbox}\fi}
           \newcommand{\fwsmall}{\columnwidth}
\newenvironment{fwfig}{\begin{figure*}[!t]}{\end{figure*}}   \newcommand{\appref}[1]{}                     \newcommand{\Appref}[1]{}
\newcommand{\efstudytitle}{Failure Study on Engine Power Loss}
\newenvironment{tbl}{\begin{table}[!t]\setlength{\tabcolsep}{3.5pt}}{\end{table}}
\newenvironment{tblwide}{\begin{table*}[!t]}{\end{table*}}

\begin{document}
\bstctlcite{IEEEexample:BSTcontrol}

\title{A Hybrid Framework for Explainable Aviation Risk: Causal Chain Extraction from Safety Reports with Physics-Based Failure Quantification and Interventional Analysis}

\author{Yutian~Pang, Yifang~Tian, Xinyu~Zhao, Jueming~Hu, John-Paul~Clarke, and Hans-Arno~Jacobsen,~\IEEEmembership{Fellow,~IEEE}\thanks{Y. Pang and J.-P. Clarke are with the Department of Aerospace Engineering and Engineering Mechanics, The University of Texas at Austin, Austin, TX 78712, USA.}\thanks{Y. Tian and H.-A. Jacobsen are with the Department of Electrical and Computer Engineering, University of Toronto, Toronto, ON M5S 3G4, Canada.}\thanks{X. Zhao is with the School of Computing and Augmented Intelligence, Arizona State University, Tempe, AZ 85287, USA.}\thanks{J. Hu is with the School of Aerospace and Mechanical Engineering, University of Oklahoma, Norman, OK 73019, USA. Corresponding author, e-mail: jueming@ou.edu.}\thanks{Y. Pang and J.-P. Clarke were supported by the NASA University Leadership Initiative program under the project ``Autonomous Aerial Cargo Operations at Scale'', grant No. 80NSSC21M071 to the University of Texas at Austin.}
}

\markboth{IEEE Journals}{Pang \MakeLowercase{\textit{et al.}}: A Hybrid Framework for Explainable Aviation Risk}

\maketitle

\begin{abstract}
Aviation authorities hold decades of accident reports whose narratives record how investigators reconstructed each accident. Computational work on these archives has mostly sorted reports into fixed taxonomies. In this work, we propose a hybrid framework in which a text-based corpus identifies the leading factors of a failure mode. Physics-based models then quantify how likely that failure is at a stated operating point. Ordered causal chains extracted from 56{,}202 American accident reports with a local large language model aggregate into a causation knowledge graph. On an evaluation sample of 2{,}000 reports, the chains recover 90.8\% of investigator-flagged causes. Each physical factor whose drivers the record carries receives an occurrence model calibrated to the accident population. A leaky noisy-OR fitted to the corpus combines the factors into a failure mode probability that supports intervention. For engine power loss, the carburetor icing model separates 1{,}205 icing accidents from other accidents using temperature and dewpoint alone, reaching 0.61 in area under the receiver operating characteristic curve against a density ratio ceiling of 0.64. Routine weather hours give no valid control, because flying concentrates in fair weather. On later years the physics adds 0.03 to that area for occurrence and nothing for severity, so its value lies in interpretability. On four named accidents, abduction gives a probability of necessity of 0.29 to 0.50 for the heat omission, under monotonicity. The Fast Causal Inference algorithm does not license the largest backdoor adjustment on the learned networks.
\end{abstract}


         \section{Introduction}\label{sec:intro}

Aviation authorities have collected a large archive of accident investigations. Each record pairs a free-text narrative with structured codes, namely an ordered sequence of occurrences and a set of findings that investigators flag as causes or contributing factors. The narratives contain causal chains such as induction icing that leads to engine power loss, then to a forced landing, and finally to terrain impact. Recovered at scale, those chains would give safety analysts a network of canonical safety factors in which each directed edge states that one factor caused another. Such a network supports the question that matters for prevention \citep{pearl2009causality}, namely how much the risk of a severe outcome would have fallen had a given contributing factor been absent.

Computational work on these archives has grown quickly, yet most of it stops short of that question in two respects. The first concerns what is extracted. A large body of work sorts narratives into fixed taxonomies with domain-adapted or prompted language models \citep{andrade2023safeaerobert,jing2023aviationbert,liu2025hfacscot}. A document label records neither the order of events nor the factor the investigator held responsible. Finer studies still classify onto code sets \citep{zhao2025hierarchical} or work on a few thousand reports \citep{siddeshwar2024towards}. The second concerns what is done with the extracted structure. Knowledge graphs built with large language models \citep{chen2024aviationkg} and Bayesian networks fitted to coded data \citep{zhang2021bayesian} quantify risk as a frequency over the reports. That frequency inherits the selection bias of an archive that records only reportable events. It is one scalar per factor with no mechanism behind it. Engineering practice models those mechanisms through fault trees \citep{sae2004arp4761}, gust load formulas \citep{pratt1954gust} and flight simulation \citep{berndt2004jsbsim}. On its own, however, it cannot say which failure modes dominate a fleet or how often each arises. Hybrid causal risk models join the two, but experts build their structure and they quantify one accident at a time. No aviation study learns that structure from a complete report archive and attaches mechanistic models to its factors.

The main contributions of this paper are the following.

\begin{itemize}
\item We recover causal structure from a complete corpus of 56{,}202 NTSB and FAA reports rather than from a curated sample. Extraction returns one ordered causal chain per narrative under a closed factor vocabulary, so each chain is acyclic by construction. Chains from different accidents then aggregate into the edge counts that a statistical causal analysis needs (Section~\ref{sec:data}).

\item We attach a physics occurrence model to every factor whose drivers are physical. Where the record carries those drivers, the model is calibrated so that physics supplies the conditional shape of the risk while the corpus supplies its absolute level (Sections~\ref{sec:physics} and \ref{sec:fusion}). Such a factor carries a risk that responds to its physical drivers, here temperature and dewpoint for carburetor icing, in place of one frequency for every flight.

\item We give a rule that combines contributing factor risks into a failure mode probability conditioned on the operating point. A single do-operator semantics covers both backdoor adjustment on the learned graph and intervention on a physical driver. A counterfactual computed by abduction applies the same models to named accidents (Sections~\ref{sec:combine} and \ref{sec:cf}).

\item We test the framework on engine power loss (Section~\ref{sec:efcase}) and develop carburetor icing as its worked physics factor.

\item We evaluate every layer of the framework and state what the archives cannot support (Section~\ref{sec:val}). The physics models fit no parameter to the corpus. The combination and the calibration links are fitted, and every fitted arm is scored out of fold or on later years. Several results are negative, and we report them as findings.
\end{itemize}

\section{Related Work}\label{sec:related}

Text mining of safety narratives began with domain-adapted language models used as classifiers, including SafeAeroBERT and Aviation-BERT \citep{andrade2023safeaerobert,chandra2023aviationbert}. It continued with large language models for human factors coding \citep{liu2025hfacscot,drones2025hfacs} and fine-level event extraction \citep{zhao2025hierarchical,siddeshwar2024towards}. Knowledge graphs assembled from extracted content map interdependencies and do not support interventional queries \citep{chen2024aviationkg,safetyscience2026kg}. The Bayesian network of \citet{zhang2021bayesian} is learned from manually coded reports with priors from departure statistics, and it conditions on evidence without intervening on the graph.

Causal inference on accident data is concentrated on the lower two rungs of Pearl's ladder \citep{pearl2009causality,gansch2025causal,karwa2011causal}. Counterfactual work that is formally sound runs on models built by experts or on a handful of cases \citep{ruiztagle2022counterfactual,zibaei2024building,scanlon2021waymo}. Language models exploit variable names and mention order and remain unreliable on whole causal graphs \citep{kiciman2023causal,jin2023cladder}, which motivates letting them supply local ordering and typed factors while the structure and probabilities come from aggregation \citep{vashishtha2025causal,zhang2024leveraging}.

Physics-based risk modeling quantifies failure from mechanism, through fault and event trees \citep{sae2004arp4761}, gust loads and continuous turbulence \citep{pratt1954gust,milf8785c,hoblit1988gust}, open flight dynamics and performance models \citep{berndt2004jsbsim,sun2020openap,nuic2010bada} and operational charts \citep{faa2009carbice}. Hybrid automata give these mechanisms a formal home \citep{tomlin2003computational,yang2022framework}. Hybrid causal risk models already combine mechanisms with accident data. Examples are hybrid causal logic \citep{groth2010hybrid}, the causal model for air transport safety and its analysis of flight TK 1951 \citep{ale2006towards,ale2009further,ale2010tk1951}, dynamic probabilistic risk assessment \citep{aldemir2013survey} and physics-informed Bayesian networks in adjacent domains \citep{meng2022datadriven,xing2025physics}. They take their structure from experts and their frequencies from exposure-normalized counts, and they quantify one accident at a time.
\subsection{Research Gap}\label{sec:rw_gap}

Corpus-driven work leaves each factor as one unconditional frequency with no mechanism behind it, so an intervention on a factor can only be a statistical adjustment. Sound counterfactual work and the hybrid causal models run on structure built by experts and quantify one accident at a time. To our knowledge, no aviation study extracts causal chains over a complete corpus, attaches physics occurrence models to the factors whose drivers are physical, and checks each reconstructed chain against the physics of the individual case.

\section{Methodology}\label{sec:method}

\subsection{Overview and Division of Labor}\label{sec:overview}

The framework has two layers joined by a fusion rule, and Fig.~\ref{fig:framework} traces the flow between them. The data layer (Section~\ref{sec:data}) turns raw reports into a causation graph and reads off, for each failure mode, its leading factors, its downstream consequences and its severity. The physics layer (Section~\ref{sec:physics}) attaches an occurrence model to any failure mode with a physical mechanism, mapping continuous physical drivers to the probability that the failure occurs. The fusion step (Section~\ref{sec:fusion}) combines the physics occurrence probability with the consequence terms from the data, and calibrates the absolute level of the physical model against the corpus. Section~\ref{sec:automaton} states the object that joins the two layers. The causal chain is a path over the modes of a hybrid automaton. Its transitions carry the corpus conditionals, and its guards are per-case physics margins that check physical admissibility before any probability is attached.

\begin{fwfig}
\centering
\includegraphics[width=\textwidth]{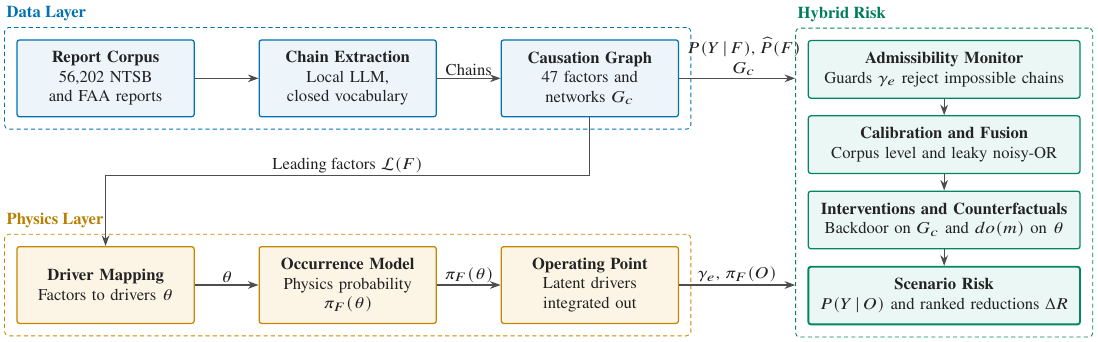}
\caption{The hybrid framework. The data layer converts 56{,}202 report narratives into ordered causal chains and aggregates them into a causation graph. The graph gives the leading factors $\mathcal{L}(F)$, the consequence conditionals $P(Y\mid F)$, the corpus base rates $\widehat P(F)$ and the per-category networks $G_c$. The physics layer maps the leading factors that have physical drivers to continuous variables $\theta$ and evaluates an occurrence model $\pi_F$ at the stated operating point. The admissibility monitor rejects any chain whose transitions the guard margins $\gamma_e$ rule out at the recorded state. Calibration ties the level of the physical model to the corpus marginal, and a leaky noisy-OR fitted to the corpus combines the contributing factors of a failure mode. Interventional queries act through backdoor adjustment on $G_c$ or through an intervention $do(m)$ that transforms the physical drivers.}
\label{fig:framework}
\end{fwfig}

Not every node receives a physical model. A node qualifies when its data-learned leading factors correspond to continuous physical variables. Human and organizational factors such as inappropriate decision or procedure not followed have no such correspondence, so they remain data only. The partition of nodes into physics-modeled and data-only nodes therefore follows from the structure of the problem rather than from an analyst's choice.

\subsection{Data Layer of Causal Chains and Leading Factors}\label{sec:data}

\paragraph{Corpus} The corpus draws on three public archives, the NTSB aviation databases \citep{ntsb_avdata} with 43{,}604 reports, the FAA accident and incident data system \citep{faa_aids} with 12{,}585 and the NTSB published investigation reports \citep{ntsb_reports} with 13.{} It holds 56{,}202 reports whose narratives run between 200 and 2{,}000 words. Every source is free to obtain, and the download code is released with this work.

Every record carries structured supervision from its source database. A cause statement accompanies 84\% of the records and a coded event sequence 47\%, and 70\%, all in the NTSB portion, carry findings that flag each factor as a cause or a contributing factor. Occurrence codes map to the occurrence categories of the CAST and ICAO Common Taxonomy Team (CICTT) \citep{cictt2017}, and injury and damage fields resolve severity. This supervision grounds the extraction prompt and serves as the reference standard of Section~\ref{sec:extractval}.

\paragraph{Chains} Qwen3.6-35B-A3B \citep{qwen2025technical,qwen36model}, a locally served mixture-of-experts model with 35 billion total and 3 billion active parameters running on vLLM \citep{kwon2023vllm}, converts each narrative into a temporally ordered chain of factor nodes. The nodes come from a closed vocabulary of 57 values, namely 35 event factors, 12 condition factors, 9 outcome types and an unknown placeholder. Every node may cite only earlier nodes as its causes, so each chain is a directed acyclic graph by construction. Formally, report $r$ yields $\mathcal{C}_r=(v_{r,1},\dots,v_{r,n_r})$ with $v_{r,k}\in\mathcal{V}$, $|\mathcal{V}|=57$, and a parent map $\mathrm{pa}(k)\subseteq\{1,\dots,k-1\}$. The run produced 55{,}940 valid chains with no schema violations, and 262 records that returned no valid or an empty chain enter no count below. We write $\mathcal{F}\subset\mathcal{V}$ for the 47 non-outcome factors and $\mathcal{Y}$ for the 9 outcomes.

\paragraph{Schema} The categories of the extraction schema come from established taxonomies. The condition, event and outcome classes follow the separation of latent conditions from active failures \citep{reason1990human,rasmussen1997risk}, the event groups follow the layers of HFACS \citep{shappell2000hfacs}, and the encounters, outcomes and flight phases follow CICTT \citep{cictt2017,cictt_phase2013}. Each node carries its position, class, factor type, a trigger phrase, a flight phase and an attribution role that mirrors the NTSB cause and factor flags \citep{ntsb_avdata}. Causality lives in a list of at most four parents per node. Each parent entry names an earlier node, grades the link as direct or contributing and quotes the supporting narrative span, so every edge of the aggregated graph traces back to checkable text. Outcome nodes may cause only other outcomes, a factor type repeated within a chain counts once in the aggregation, and constrained decoding against the JSON grammar enforces the schema with one repair turn on violation.

\paragraph{Aggregation} Chains aggregate at the factor level into a directed graph. For factors $i,j\in\mathcal{F}$, let $s_i$ be the number of reports whose chain contains factor $i$. The edge support $s_{ij}$ counts the reports that contain at least one causal link from a node of type $i$ to a node of type $j$. Each edge then carries
\begin{equation}
P(j\mid i)=\frac{s_{ij}}{s_i},\qquad
\mathrm{lift}(i,j)=\frac{P(j\mid i)}{P(j)}=\frac{s_{ij}/s_i}{s_j/N},
\label{eq:edge}
\end{equation}
where $N$ is the number of reports. Aggregating 55{,}940 chains gives a compact graph of 47 factors and 434 edges with support of at least 30. The edge tables of the failure study{} read their statistics from the unfiltered edge list. Each report also contributes a binary presence vector $\mathbf{x}_r\in\{0,1\}^{47}$ and a resolved severity $Y_r$, where structured injury and damage fields override the model's own estimate. These vectors form the table used in the statistical analyses below.

\paragraph{Leading Factors} The leading factors of a failure mode $F\in\mathcal{F}$ are the antecedents that the corpus attributes to it,
\begin{equation}
\mathcal{L}(F)=\bigl\{\, i\in\mathcal{F}:\; s_{iF}\geq \rho\, s_F \,\bigr\},
\label{eq:leading}
\end{equation}
the direct parents whose edge into $F$ explains at least a fraction $\rho$ of the reports that carry $F$, with $\rho=0.02$ held fixed across failure modes. No lift threshold enters, because lift within an accident population is subject to the selection effect described next. The failure study reports{} lift as a diagnostic only. The set $\mathcal{L}(F)$ is the answer of the data layer to the question of what leads to $F$. We identify physical drivers within this set, and we pose interventions on its members.

\paragraph{Selection on the Accident} The corpus observes a flight only when it ended in a reportable event. Conditioning on that event makes the causes compete for the same accident. The presence of one sufficient cause then lowers the chance that another is recorded in the same report, which is a collider effect of the selection \citep{hernan2004structural}. Its signature is a lift below one on an edge that is causal in the flight population. Procedure not followed enters engine failure at lift 0.24, inadequate maintenance at 0.39 and structural failure at 0.42 (Table~\ref{tbl:effactors}). The fitted combination of Section~\ref{sec:combine} gives structural failure no direct weight and keeps strengths of 0.11 and 0.35 for the other two, so a low lift alone does not decide the strength.{} Within-accident contrasts therefore measure competition among causes as much as causation, and the causal sufficiency that the PC algorithm assumes fails by construction. Section~\ref{sec:backdoorres} runs FCI, which allows latent confounding and selection, alongside PC on the same data.

\paragraph{Structure per Category} For a directed analysis that accounts for confounding, we group accidents by occurrence category, read from the NTSB occurrence codes (Section~\ref{sec:backdoorres}), and learn a causal network $G_c$ per category from $\{\mathbf{x}_r\}$. The PC algorithm \citep{spirtes2000causation,zheng2024causallearn} provides the skeleton under $\chi^2$ independence tests. Ordering priors from a language model, in the manner of \citet{vashishtha2025causal} and \citet{zhang2024leveraging}, and empirical chain precedence then orient the edges, and bootstrap resampling filters them. The network $G_c$ supplies the adjustment sets used in Section~\ref{sec:cf}.

\subsection{Physics Layer of Failure Occurrence Models}\label{sec:physics}

Let $F$ be a failure mode with a physical mechanism and continuous physical drivers $\theta\in\Theta_F$. The physics layer supplies an occurrence model
\begin{equation}
\pi_F(\theta)\;=\;\Pr(F\text{ occurs}\mid\theta)\;=\;g_F(\theta;\phi_F),
\label{eq:pi}
\end{equation}
where $g_F$ follows from the governing physics and $\phi_F$ collects the model parameters. The drivers $\theta$ are the physical realization of the data-learned leading factors $\mathcal{L}(F)$. A qualitative antecedent such as moist air becomes the recorded dewpoint, and low power becomes a power setting.

In general, only a subset $O$ of the components of $\theta$ is recorded for a given accident. The latent remainder $\xi$ is marginalized over a prior conditioned on flight phase,
\begin{equation}
\pi_F(O)=\int_{\Xi}\mathbf{1}\{F\mid O,\xi\}\,p(\xi\mid O)\,d\xi.
\label{eq:marginalize}
\end{equation}
No factor in this paper requires that step. Carburetor icing reads temperature and dewpoint from the record and is evaluated at cruise power and at descent power separately.
The corpus enters Eq.~\eqref{eq:pi} in three ways. The choice of $g_F$ follows from the mechanism that the corpus identifies. The prior $p(\xi\mid O)$ takes its shape from the observed distribution of power settings, phases and aircraft classes in accidents of that mode. The parameters $\phi_F$ are anchored to the corpus by the calibration described next.

\subsection{The Chain Automaton and Admissibility Monitoring}\label{sec:automaton}

A causal chain is a walk through discrete failure states, and each occurrence model of Section~\ref{sec:physics} decides from the continuous state of a flight whether a step was physically available. We take the hybrid automaton \citep{tomlin2003computational,yang2022framework}, keep its modes and guards and drop the flows, which a narrative rarely supports. The modes are the factors and outcomes of Section~\ref{sec:data}, and the transitions are the edges of the causation graph. A guarded transition $e$ carries a guard margin $\gamma_e(\theta)$, a signed quantity with physical units whose sign is oriented so that $e$ is enabled exactly when $\gamma_e(\theta)\ge 0$. Table~\ref{tbl:guards} lists the guards of the failure study. Their drivers are the elapsed or planned flight time $t$ against the fuel endurance $t_{\mathrm{end}}$, and the condensable water $w$ of Eq.~\eqref{eq:icewater} against its threshold $w_{\mathrm{thr}}$. Each is a quantity its occurrence model already computes, so the guards add no new physics. A transition is checkable for a case when it carries a guard and the record carries the drivers of that guard. The admissibility monitor rejects a chain when a checkable transition has $\gamma_e(O)<0$ at the recorded observables $O$. The chain then requires a step the physics rules out, so either the extraction is wrong or a mode is missing. No reweighting of a probabilistic blend reproduces such a rejection \citep{kim2025guaranteed}. The verdict is invariant under any strictly increasing transformation of $\gamma_e$, so it needs no calibration. Two corpus statistics follow. The guard coverage is the share of transitions that are checkable, and the violation rate is the share of checkable transitions that the monitor rejects. A corrupted guard control, with sign-inverted margins or the weather of an arbitrary accident, separates extraction error from model inadequacy. For a failure onset the guard condition is the occurrence event itself. Carburetor icing maps its margin through the monotone link that Eq.~\eqref{eq:calib} calibrates, so rankings across operating points come from the guard alone. The monitor checks single transitions at recorded states, not reachability over continuous trajectories.

\begin{tbl}
\centering
\caption{Guard margins behind the physics-modeled transitions{} of the failure study.}
\label{tbl:guards}
\begin{tabular}{@{}lll@{}}
\toprule
Transition into & Guard margin $\gamma_e(\theta)$ & Units \\
\midrule
Fuel exhaustion & $t-t_{\mathrm{end}}$ & hours \\
Carburetor icing & $w(T,T_d,\varpi)-w_{\mathrm{thr}}$ & grams per cubic meter \\
\bottomrule
\end{tabular}
\end{tbl}

\subsection{Fusion and Tiered Calibration}\label{sec:fusion}

The monitor decides which transitions are available for a case, and the fusion layer attaches probabilities to the available ones. Each node draws its occurrence probability from the best available source, organized as a tier. Tier~1 is a physics model, Eq.~\eqref{eq:pi}. Tier~2 is published failure rate data. Tier~3 is the corpus marginal $\widehat P(F)$. Tier~4 is expert or model elicitation. Tier~3 forms the backbone, because it is defined for every factor in the corpus, while Tier~1 sharpens nodes with a mechanism.

Where a physics estimate and a corpus estimate both exist, they occupy different probability spaces. The corpus frequency is conditioned on a reportable accident having occurred, whereas $\pi_F$ is a per-exposure probability with no population behind it. We therefore treat physics as the source of the conditional shape and the corpus as the source of the absolute level. Calibration fixes the level so that the physical model reproduces the corpus marginal once averaged over the observed operating point distribution $\mathcal{D}$ of the accident population,
\begin{equation}
\mathbb{E}_{O\sim\mathcal{D}}\!\left[\pi_F(O;\phi_F)\right]
\;=\;\widehat P(F).
\label{eq:calib}
\end{equation}
This step keeps the relative sensitivity that makes interventions informative and puts the physics term on the population the corpus observes. Section~\ref{sec:calibres} executes it for carburetor icing as a monotone link on the physics score. The fuel term cannot be calibrated, since fuel on board and flight time are not coded fields. Every quantity then has a stated reference class. The corpus base rates, the edge conditionals, the fitted combination of Section~\ref{sec:combine} and the calibrated physics terms refer to a reportable accident, the guard verdicts to no population, and the uncalibrated chart probability of Eq.~\eqref{eq:pice} and the fuel term to an exposure, so the failure study reports the effect of the last two as ratios.

Consider a physics-modeled failure $F$ and a downstream event or outcome $Y$ that is reachable from $F$ in the graph. The corpus supplies the consequence conditional $P(Y\mid F)$, either as a single edge from Eq.~\eqref{eq:edge} or as a product along a path. The fused risk that $F$ produces $Y$ at operating point $O$ is then
\begin{equation}
R_F(Y\mid O)\;=\;\underbrace{\pi_F(O)}_{\text{physics}}\;\cdot\;
\underbrace{P(Y\mid F)}_{\text{data}}.
\label{eq:fusion}
\end{equation}
Here the physics term states whether the failure will occur under the stated conditions, while the corpus supplies how often it goes on to produce $Y$. Equation~\eqref{eq:fusion} quantifies a chain of depth two, one guarded onset followed by one consequence transition.{} The failure study quantifies{} chains of this depth, which keeps every reported number attributable to one named mechanism and one corpus conditional. Uncertainty propagates by carrying tier-specific distributions on both terms and sampling from them, which yields a confidence interval on $R_F$.

\subsection{Combining Contributing Factors}\label{sec:combine}

Equation~\eqref{eq:fusion} quantifies a single failure. A higher-level failure event $Y$, such as engine failure, is reached through several contributing factors. The data layer identifies them as its causal parents $\mathrm{pa}(Y)=\{X_1,\dots,X_k\}$, the leading factor set of Eq.~\eqref{eq:leading}. Each contributing factor carries an occurrence probability $\pi_i$, taken from its calibrated physics model where one exists and otherwise from the corpus base rate $\widehat P(X_i)$. It also carries a strength $p_i$, the probability that the factor alone produces $Y$. We treat the contributing factors as independent causes of the top event, as in a noisy-OR or bow-tie model, with a leak $\ell$ for the causes the set omits. The failure mode probability then becomes
\begin{equation}
P(Y\mid O)=1-(1-\ell)\prod_{i=1}^{k}\bigl(1-\pi_i(O)\,p_i\bigr),
\label{eq:combine}
\end{equation}
and the contribution of factor $i$ is the drop that follows when the factor is removed,
\begin{equation}
\Delta_i = P(Y\mid O)-P\bigl(Y\mid O,\;\pi_i{=}0\bigr).
\label{eq:contribution}
\end{equation}
We fit the leak and the strengths by maximum likelihood on the binary presence table of Section~\ref{sec:data}, with one row per report. The strength $p_i$ therefore carries the noisy-OR meaning of a cause acting alone. It differs from the marginal conditional $q_i=P(Y\mid X_i)$ of Eq.~\eqref{eq:edge}, which counts the work of the other parents as well. The leak absorbs the causes that the leading factor set does not list, and the mean of the fitted probabilities over the records reproduces the prevalence of $Y$. Evaluated at the corpus base rates, Eq.~\eqref{eq:combine} gives instead the population value, which differs slightly because the parents are not independent in the presence table. A parent whose within-accident lift is below one often receives a strength near zero, for one of two reasons. Either its effect runs through a parent already in the set, or its association with $Y$ reflects the selection on the accident described in Section~\ref{sec:data}. Equation~\eqref{eq:contribution} corresponds to an intervention that removes $X_i$ at the level of the failure mode. The physics-modeled factors supply $\pi_i(O)$ as a function of the operating point, so $P(Y\mid O)$ depends on the scenario. The data-only factors enter at their corpus base rates. Equation~\eqref{eq:combine} is the one-level combination, with parent activations $\pi_i(O)$ and strengths $p_i$ entering unmonitored where no consequence guard applies. It gives an interpretable structural model.

\subsection{Interventional and Counterfactual Analysis}\label{sec:cf}

The framework answers interventional queries through the do-operator in two forms. The form depends on whether the target of the intervention is a categorical factor or a physical driver. The framework also answers one counterfactual query by abduction on the guarded mechanisms.

\paragraph{Graph-Level Intervention} Consider a categorical factor $X\in\mathcal{L}(F)$ that has no physical model, and a binary severe outcome variable $Y$ in category $c$. Comparing accidents with and without $X$ is confounded, because $X$ occurs together with its own causes. We therefore apply backdoor adjustment on the learned network and stratify on the parents of $X$, written $Z=\mathrm{pa}_{G_c}(X)$,
\begin{equation}
P(Y\mid do(X{=}x))=\sum_{z}P(Y\mid X{=}x,Z{=}z)\,P(Z{=}z),
\label{eq:backdoor}
\end{equation}
with Laplace-smoothed stratum estimates and effect $\tau_X=P(Y\mid do(X{=}1))-P(Y\mid do(X{=}0))$.

\paragraph{Physics-Level Intervention} For a failure mode that carries a physics model, an intervention acts on a physical driver, or on a mitigation $m$ that maps to a driver transform $\mathcal{T}_m:\theta\mapsto\theta'$. We then reevaluate the occurrence model at the intervened value,
\begin{equation}
\begin{split}
R_F\!\left(Y\mid do(m),O\right)&=\pi_F\!\left(\mathcal{T}_m(\theta)\right)\cdot
P(Y\mid F),\\
\Delta R &= R_F(Y\mid O)-R_F(Y\mid do(m),O).
\end{split}
\label{eq:phys_do}
\end{equation}
The intervention propagates through a mechanistic model, so $\Delta R$ is attributable to a named physical cause and monotone in the driver, and the monitor reevaluates $\gamma_e(\mathcal{T}_m(\theta))$. Equations~\eqref{eq:backdoor} and \eqref{eq:phys_do} estimate the same reduction in severe outcome probability, statistically where only categorical evidence exists and mechanistically where a physical model applies.

\paragraph{Counterfactual on a Named Accident} Both forms are interventional, the second rung of Pearl's ladder. A counterfactual, the third rung, asks what would have happened in a particular accident had one of its causes been different. It needs abduction, action and prediction on a structural model with explicit noise. The guarded mechanisms of Section~\ref{sec:automaton} supply such a model. For carburetor icing the structural equation is $\mathrm{ICE}=\mathbf{1}\{U\le g(w)\}$, with $w$ the condensable water at the recorded temperature and dewpoint, $g$ the calibrated link of Eq.~\eqref{eq:calib} and $U$ uniform on $[0,1]$. Abduction restricts $U$ to the values consistent with what the record shows. The action $do(\mathrm{heat})$ warms the intake so that the condensable water becomes $w'$, and prediction pushes the same $U$ through the modified equation. Under this monotone coupling the probability that icing would not have occurred had heat been applied, given that it did occur, is
\begin{equation}
\mathrm{PN}_{\mathrm{ice}}=1-\frac{g(w')}{g(w)},
\label{eq:pn}
\end{equation}
the probability of necessity of the heat omission \citep{tian2000probabilities}. The engine failure node is the leaky noisy-OR of Eq.~\eqref{eq:combine}, so the same three steps give the probability of necessity for the power loss. The other fitted parents enter at their base rates and stand in for the causes the record does not name. Section~\ref{sec:pn} reports the quantity on four named accidents together with the bounds that hold without the monotonicity assumption.

\section{\efstudytitle}\label{sec:efcase}

The failure study applies the machinery of Section~\ref{sec:method} to engine failure, the most frequent failure node of the causation graph with 16{,}406 records. It develops carburetor icing, the dominant weather-driven parent of engine failure, end to end.

\subsection{Contributing Factors}\label{sec:effactors}

Table~\ref{tbl:effactors}{} lists the nine leading factors of Eq.~\eqref{eq:leading}, the parents with edge support of at least 329. The table also gives their base rates, occurrence models, fitted strengths $p_i$ and contributions $\Delta_i$, with 95\% bootstrap intervals over 500 refits of the combination. Its entries are rounded, while the text computes from unrounded values.{} Engine failure suits the hybrid treatment well, because its dominant contributing factors are fuel state and induction mechanisms that carry the strongest causal edges in the graph. Fuel exhaustion or starvation leads with support 5{,}051, $P(\text{EF}\mid\text{fuel})=0.93$ and lift 3.18, which is the strongest single edge in the causation graph. Fuel contamination follows at lift 2.57 and carburetor icing at lift 2.27. The set covers most chains. Of the 16{,}406 records with an engine failure node, 77\% give that node a member of the set as an explicit parent, 6\% give it only parents outside the set and 17\% give it no parent. The leak of Section~\ref{sec:efcombine} therefore stands mostly for power losses that the narrative states without a cause. Table~\ref{tbl:effactors} feeds the combination rule.{} Structural failure is a parent because, in about nine of ten records behind this edge, it names a fractured powerplant part, such as a propeller blade, cylinder head or connecting rod. Section~\ref{sec:backdoorres} adjusts its effect on a severe outcome within the airframe failure category, which supplies no adjustment set to the combination of Section~\ref{sec:efcombine}.

\begin{tblwide}
\centering
\caption{Leading factors of engine failure under the rule of Eq.~\eqref{eq:leading}.}
\label{tbl:effactors}
\fittab{\begin{tabular}{@{}lrrrrlrrl@{}}
\toprule
Factor & Edge support & $q_i$ & Lift & Base rate & Model & $p_i$ & $\Delta_i$ & 95\% CI \\
\midrule
Fuel exhaustion or starvation & 5{,}051 & 0.93 & 3.18 & 0.097 & physics (endurance)$^{\dagger}$ & 0.96 & 0.071 & (0.070, 0.073) \\
Latent mechanical defect & 2{,}331 & 0.28 & 0.95 & 0.149 & data & 0.19 & 0.021 & (0.019, 0.022) \\
Procedure not followed & 1{,}137 & 0.07 & 0.24 & 0.293 & data & 0.11 & 0.023 & (0.021, 0.025) \\
Other system failure & 1{,}035 & 0.28 & 0.97 & 0.065 & data & 0.20 & 0.009 & (0.008, 0.010) \\
Carburetor or induction icing & 971 & 0.67 & 2.27 & 0.026 & physics (icing) & 0.93 & 0.017 & (0.016, 0.018) \\
Fuel system anomaly & 954 & 0.57 & 1.94 & 0.030 & surrogate & 0.63 & 0.013 & (0.012, 0.014) \\
Fuel contamination & 826 & 0.75 & 2.57 & 0.020 & surrogate & 0.92 & 0.013 & (0.012, 0.014) \\
Inadequate maintenance & 649 & 0.12 & 0.39 & 0.101 & data & 0.35 & 0.025 & (0.024, 0.027) \\
Structural failure & 333 & 0.12 & 0.42 & 0.049 & data& 0.00 & 0.000 & (0.000, 0.001) \\
\bottomrule
\end{tabular}}
\par\smallskip{\footnotesize $^{\dagger}$The endurance model is not calibrated, since fuel on board and flight time are not coded fields. The factor enters the combination at its corpus base rate.\par}
\end{tblwide}

\subsection{Physics of Fuel Exhaustion}\label{sec:fuelphys}

The dominant factor has transparent physics, but the combination uses its model only for one ratio. Exhaustion occurs when the flight outlasts the endurance $E=m_f/\dot m$ of the usable fuel mass $m_f$ at fuel flow $\dot m$. With reserve $r=E-t_{\mathrm{flight}}$ and a Gaussian margin of scale $\sigma=c\,t_{\mathrm{flight}}$ for the uncertainty in fuel flow and routing, the exhaustion probability is $\pi_{\mathrm{fuel}}(r)=\Phi(-r/\sigma)$. With $c=0.25$ on a 3 h flight it is 0.25 at a reserve of 0.5 h and 0.004 at 2 h. Fuel on board and flight time are not coded fields, so Eq.~\eqref{eq:calib} cannot calibrate the term, and the combination takes this factor at its corpus base rate. The remaining factors use the carburetor icing physics of Section~\ref{sec:icephys}, component reliability surrogates for the fuel system and contamination, and corpus rates for defect, maintenance and procedure. {} Standard performance models \citep{sun2020openap,nuic2010bada} supply the glide and forced landing consequences downstream of the failure.

\subsection{Carburetor Icing as the Worked Physics Factor}\label{sec:icing}

We now develop one contributing factor end to end. Carburetor icing is the dominant weather-driven parent of engine failure, and its edge into engine failure carries support 971 in the graph. It suits the role of worked physics factor for three reasons. Its physical drivers are recorded for nearly every report, and the failure is a published function of those drivers. The power setting, the one driver the record lacks, can be bracketed by evaluating the model at cruise and at descent power. A stall, by contrast, depends on airspeed and load factor that no record keeps.

\subsubsection{Physics of Induction Icing}\label{sec:icephys}

In a float carburetor, the static pressure drop across the venturi and the vaporization of fuel can lower the temperature substantially where the inducted air and fuel mix before combustion. If the mixture cools below its dewpoint and below 0~$^\circ$C, moisture condenses and freezes on the throttle plate, which restricts the induction airflow and reduces power. Let $T$ and $T_d$ be the ambient temperature and dewpoint in degrees Celsius. Saturation vapor pressure follows the Magnus form \citep{alduchov1996magnus},
\begin{equation}
e_s(T)=6.112\,\exp\!\left(\frac{17.62\,T}{243.12+T}\right)\ \text{hPa},
\qquad \mathrm{RH}=\frac{e_s(T_d)}{e_s(T)}.
\label{eq:magnus}
\end{equation}
The mixture cools to a throttle plate temperature $T_{\mathrm{thr}}=T-\Delta T(\varpi)$. The cooling $\Delta T$ depends on the power setting $\varpi$ and is largest at low power, because a nearly closed throttle produces the greatest pressure drop. The carburetor icing probability chart of FAA Special Airworthiness Information Bulletin CE-09-35 \citep{faa2009carbice} carries separate zones for serious icing at descent power for this reason. The chart attaches no percentage of rated power to its settings, and these settings vary with the aircraft type and the pilot. The model therefore represents each setting by its cooling alone, 25~$^\circ$C at cruise power and 32~$^\circ$C at the reduced power of a descent or glide. The moisture that condenses in the subzero region, and therefore freezes, is proportional to the vapor pressure difference
\begin{equation}
\begin{split}
w(T,T_d,\varpi)=\max~\!\bigl(0,\;e_s(\min(T_d,0))&-e_s(T_{\mathrm{thr}})\bigr)\\
&\cdot\mathbf{1}\{T_{\mathrm{thr}}<0\},
\end{split}
\label{eq:icewater}
\end{equation}
and the occurrence probability is a saturating exposure function of that quantity,
\begin{equation}
\pi_{\mathrm{ice}}(T,T_d,\varpi)=1-\exp~\!\bigl(-\kappa\,w(T,T_d,\varpi)\bigr).
\label{eq:pice}
\end{equation}
The quantity $w$ carries pressure units, so $\kappa$ is an exposure constant per hectopascal that absorbs the induction mass flow and the exposure time. The guard margin of Table~\ref{tbl:guards} converts the same difference to an absolute humidity at the throttle plate temperature, $216.68\,(w-w_{\mathrm{thr}})/(T_{\mathrm{thr}}+273.15)$ in grams per cubic meter. Here $w_{\mathrm{thr}}$ is the onset of the light zone of the chart. The two parameters $\phi=\{\Delta T(\varpi),\kappa\}$ are the free constants of the physical model. We set them so that Eq.~\eqref{eq:pice} reproduces the named zones of that chart. This gives $\kappa=0.60$ per hectopascal and $w_{\mathrm{thr}}=0.176$~hPa, together with the two cooling values above. The chart, and not the corpus, is therefore the empirical content behind these two constants. The calibration of Eq.~\eqref{eq:calib} is a separate step that Section~\ref{sec:calibres} executes on the accident population. The resulting risk surface (Fig.~\ref{fig:icing}) recovers the less obvious features of the chart, including serious icing at warm ambient temperatures when humidity is high.

\subsubsection{Data Support}\label{sec:icedata}

The physical drivers are recorded for 79.9\% of the 93{,}360 NTSB event records once missing readings coded as zero and impossible values are discarded. Carburetor icing is a well supported cause of engine failure, with edge support 971. Carburetor icing accidents carry $P(\text{serious or fatal})=0.23$, below the corpus baseline of 0.26, because induction icing usually ends in survivable forced landings.

\begin{figure}[htbp]
\centering
\includegraphics[width=\fwsmall]{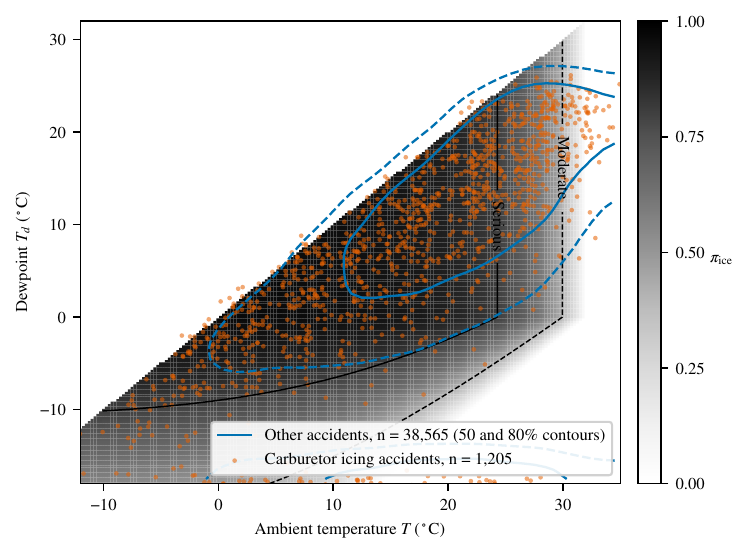}
\caption{Physics occurrence surface $\pi_{\mathrm{ice}}(T,T_d)$ at descent power, from Eqs.~\eqref{eq:magnus} to \eqref{eq:pice}. Dashed and solid contours mark the moderate value of 0.40 and the serious value of 0.80 from the published induction icing chart. The overlaid points are the 1{,}205 NTSB accidents that the narratives attribute to carburetor icing. The blue contours enclose 50 and 80\% of the 38{,}565 other weather-recorded accidents, so the figure shows separation rather than concentration alone. The AUC between the two groups is 0.61.}
\label{fig:icing}
\end{figure}

To test the physics against the data, we joined the model to the 39{,}770 extracted NTSB accidents with usable weather and asked whether the 1{,}205 that the narratives attribute to carburetor icing fall in higher-risk weather. They do. At descent power the physics assigns them a mean occurrence probability of 0.76 against 0.63 for the other accidents, and 88.9\% of them fall in the moderate or serious zone against 75.0\%. As a score, $\pi_{\mathrm{ice}}$ separates the two groups with an area under the receiver operating characteristic curve (AUC) of 0.61 (Fig.~\ref{fig:icing}), close to the density ratio ceiling of 0.64 that Section~\ref{sec:ceiling} estimates. Conducive weather is a necessary condition, and whether ice brings the aircraft down depends on carburetor heat use and engine susceptibility, whose realized rate the corpus edge $P(\text{engine failure}\mid\text{icing})=0.67$ supplies.

\subsubsection{Causal Chain to Icing}\label{sec:icechain}

The causal edges of the carburetor icing node show two features. The consequence is pinned down sharply, since the edge to engine failure carries support 971 with $P(\text{engine failure}\mid\text{icing})=0.67$ and lift 2.27. The antecedents are sparse and weak. Only procedure not followed, which in these chains records carburetor heat applied late or not at all, meets the rule of Eq.~\eqref{eq:leading}. In causal terms carburetor icing behaves as a root condition. Its true antecedents are the continuous physical drivers $\theta=(T,T_d,\varpi)$, which are not nodes in the categorical vocabulary. The occurrence model $\pi_{\mathrm{ice}}$ supplies the missing antecedent quantification, so carburetor icing enters the engine failure combination of Section~\ref{sec:efcombine} as a calibrated physics parent.

\subsubsection{Intervention Results}\label{sec:icecf}

Consider a representative carburetor icing scenario at $T=13$~$^\circ$C and $T_d=12$~$^\circ$C at descent power, warm and humid air that poses no airframe icing hazard. The chart physics gives $\pi_{\mathrm{ice}}=0.94$, in the serious zone of the chart. The calibrated link of Section~\ref{sec:calibres} maps that operating point to 0.0377 on the corpus population, against a base rate of 0.026. The fused risk of a power loss induced by carburetor ice is then the product of 0.0377 and the corpus edge 0.67, or $R_{\mathrm{ice}}=0.025$. Two physics-level interventions follow from Eq.~\eqref{eq:phys_do}. Carburetor heat is modeled as restoring 30~$^\circ$C to the intake, so that $T_{\mathrm{thr}}$ rises above freezing. It brings the modeled condensable water to zero and takes the calibrated term to 0.0079, the floor of the link, so $R_{\mathrm{ice}}$ falls by a factor of 4.8. The model applies heat before ice forms, and late or insufficient heat leaves part of the risk in place, as the coldest case of Section~\ref{sec:pn} shows. A drier atmosphere, with $T_d$ falling from 12 to $-5$~$^\circ$C, attenuates the risk without removing it, taking the calibrated term to 0.0275 and $R_{\mathrm{ice}}$ down by 27\%. Each reduction traces to a named driver, and the carburetor heat case is directly actionable in flight.

\subsubsection{Probability of Necessity on Named Accidents}\label{sec:pn}

Table~\ref{tbl:pn} applies Eq.~\eqref{eq:pn} to four accidents whose chains carry carburetor icing, a delayed or omitted application of carburetor heat, and a power loss. Its link values $g(w)$ and $g(w')$ are the calibrated terms at the recorded weather and with the intake warmed by 30~$^\circ$C. In each case the NTSB probable cause names carburetor icing or the late use of carburetor heat, and the investigators judged the weather conducive to it. Given that carburetor icing occurred, the probability that it would not have occurred under carburetor heat is 0.71 in the three warmer cases and 0.41 in the coldest. In the warmer cases heat removes the condensable water, and every operating point without it takes one value of the link, the floor of 0.008, which is the carburetor icing share of the 6{,}535 weather-recorded accidents with no condensable water. Their recorded weather also falls on one step of the link, whose top is nearly flat across the moderate and serious zones. The three cases therefore receive the same value. In 20141208X13517, at $-12.2$~$^\circ$C, the warmed throttle plate stays below freezing and the chart term falls only from 0.73 to 0.14. The counterfactual therefore separates the cases by whether heat clears the ice, while the link cannot separate their weather. For the power loss, with carburetor icing observed and the other fitted parents of engine failure at their base rates, the probability of necessity is 0.50 in the warmer cases and 0.29 in the coldest. Taking only the weather and the power loss as evidence lowers it to between 0.02 and 0.04, since fuel, mechanical and procedural causes compete with carburetor icing. The bounds that hold without monotonicity run from that lower value to one, so the informative content of these numbers rests entirely on the monotone coupling of Eq.~\eqref{eq:pn}. The computation is the third rung of the ladder on a named case, and it inherits the reference class of the calibrated link and the fitted combination.

\begin{tbl}
\centering
\caption{Probability of necessity of the carburetor heat omission on four named NTSB accidents, from Eq.~\eqref{eq:pn}.}
\label{tbl:pn}
\footnotesize\fittab{\begin{tabular}{@{}lrrrrrrl@{}}
\toprule
Record & $T$ ($^\circ$C) & $T_d$ ($^\circ$C) & $g(w)$ & $g(w')$ & $\mathrm{PN}_{\mathrm{ice}}$ & $\mathrm{PN}_{\mathrm{EF}}$ & Bounds \\
\midrule
20001212X19636 & 22.2 & 18.9 & 0.028 & 0.008 & 0.71 & 0.50 & (0.04, 1) \\
20141208X13517 & $-12.2$ & $-12.8$ & 0.025 & 0.015 & 0.41 & 0.29 & (0.02, 1) \\
20020225X00252 & 2.8 & $-6.1$ & 0.028 & 0.008 & 0.71 & 0.50 & (0.04, 1) \\
20200427X62554 & 13.9 & $-2.8$ & 0.028 & 0.008 & 0.71 & 0.50 & (0.04, 1) \\
\bottomrule
\end{tabular}}
\end{tbl}

\subsection{Combining the Factors}\label{sec:efcombine}

\begin{figure*}
\centering
\includegraphics[width=\textwidth]{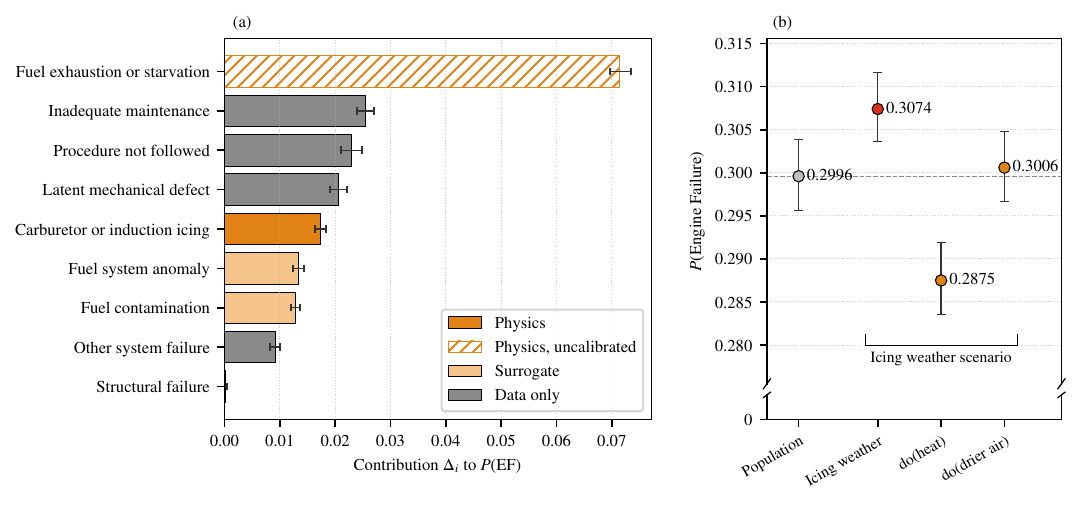}
\caption{Engine failure risk assembled from the contributing factor models. (a) Per-factor contribution to $P(\text{EF})$ by model tier, covering physics, surrogate and data-only factors. Fuel exhaustion is the dominant contributor. Its endurance model is not calibrated, so the factor enters at its corpus base rate (hatched). (b) The combined $P(\text{EF})$ conditioned on the operating point, with every term on the accident population. The vertical axis is broken between 0 and 0.28. Weather in the serious zone of the carburetor icing chart raises the population value of 0.2996 to 0.3074. The two interventions act on that scenario, carburetor heat lowering the risk to 0.2875 and drier air to 0.3006. Error bars are 95\% bootstrap intervals over 500 refits of the corpus terms, which refit the leak and the strengths on resamples of the records.{} They cover the corpus terms only.}
\label{fig:ef}
\end{figure*}

The fitted combination gives a leak of $\ell=0.084$. The mean of the fitted probabilities over the records is 0.292, against an empirical prevalence of 0.293. Evaluated at the base rates, Eq.~\eqref{eq:combine} gives the population value $P(\text{EF})=0.2996$, with a 95\% interval of $(0.2956,0.3039)$. It sits slightly above the prevalence because the product form ignores the dependence among the parents, and it is the reference for the scenarios below. Figure~\ref{fig:ef}a shows the decomposition. Fuel exhaustion dominates at 0.071 and carries a physics model, and inadequate maintenance, procedure not followed and latent defect follow at 0.021 to 0.025. Conditioning on an operating point makes the risk specific to the scenario (Fig.~\ref{fig:ef}b). Consider weather in the serious zone of the carburetor icing chart, at $T=13$~$^\circ$C and $T_d=12$~$^\circ$C at descent power, where the calibrated carburetor icing term is 0.0377. This weather raises the risk to 0.3074, a factor of 1.03. Carburetor heat, with a term of 0.0079, lowers it to 0.2875, and the drier air, with 0.0275, to 0.3006. Only the icing term changes, so each value follows from the population value and Table~\ref{tbl:effactors}. The effect is small on the accident population because an accident in such weather carries the carburetor icing factor 1.45 times as often as an average accident. The chart probability of 0.94, read as a level, would have implied 36 times. The fuel term stays out of these numbers, since it is a per-exposure probability and its strength of 0.96 was fitted on accidents. On its own scale, a reserve of 2 h on a 3 h flight gives $\pi_{\mathrm{fuel}}=0.004$ against 0.25 for a reserve of 0.5 h, and we report the intervention only as that ratio. Interventions on different parents also combine in Eq.~\eqref{eq:combine} as independent causes by construction, so the framework does not test composition beyond that property.

\section{Evaluation and Robustness}\label{sec:val}

\subsection{Fidelity of the Extracted Chains}\label{sec:extractval}

We score the extraction against the NTSB coded fields on an evaluation sample of 2{,}000 reports.{} We draw the sample from the NTSB records that carry cause and factor flags, the only per-factor reference the archives provide. A judge model matches each cause-flagged finding to the extracted nodes that state the same factor, allowing paraphrase. The judge is the same Qwen3.6-35B-A3B that performs the extraction, run with constrained JSON output and no reasoning trace. No fixed crosswalk from the vocabulary to the NTSB codes exists or is used, and the judge prompt and every match are released with the code. A domain-literate annotator, blind to the judge, repeated the matching on 50 of these reports. Annotator and judge agree on 85 of the 88 findings (96.6\%), and in all three disagreements the annotator accepted a match that the judge had rejected. Agreement over the 580 finding and node pairs is 75\%, because the annotator also accepted nodes closely related to a finding. On this sample the judge is stricter than the human reader. Recall of cause-flagged factors is 90.8\% from the narrative alone and 93.1\% when the coded findings are supplied in context. We cite the narrative-only figure for the method, since the grounded configuration, which is the production setting, sees the answer it is scored against. Of the 3{,}481 scored findings, 3{,}450 come from records before 2008, which we group by subject keywords. Recall is 0.96, 0.95, 0.96 and 0.93 for the 604 aircraft, 284 environment, 183 organizational and 1{,}581 personnel findings and 0.79 for the 798 the keywords do not classify, 3{,}133 of 3{,}450 in all.{} The reference therefore describes accidents before 2008, while the corpus runs to 2026 and drifts over that span (Section~\ref{sec:heldout}). The 90.8\% holds for the older records and is assumed for the later ones. The FAA records are extracted but not scored this way, so the measured fidelity speaks for the NTSB portion and is assumed rather than shown for the rest.

Two features of the gold standard bound these numbers. First, the reference is a set of cause flags, whereas a chain also carries intermediates and outcomes, so we score precision by role. Of the nodes marked as primary causes 62\% match a cause flag, of those marked contributing 12\%, or 41\% over both roles, and 63\% of event nodes match a coded occurrence on the 1{,}893 records that carry one. The gap follows the NTSB flags. Before 2008 contributing items carry a factor flag, which the reference omits. Against those 2{,}385 entries 39\% of contributing and 13\% of primary nodes match. Against the cause and factor flags together, the primary and contributing roles match at 69\% and 47\%, with the judge rerun in FP8 agreeing on 97.9\% of the cause flags. The all-node figure of 29.8\% mostly measures chain completeness against a sparse reference.

Second, the coded event sequence follows a coding convention within an occurrence, with the proximate subject first, and is chronological between occurrences. Agreement with it is $\tau=0.31$ over all ordered pairs and 0.82 over cross-occurrence pairs, with 83\% of reports at or above 0.75.

Two further checks support these readings. On 1{,}500 Transportation Safety Board of Canada summaries, whose coded sequence is chronological throughout, the all-pairs agreement is $\tau=0.66$. On 157 full investigation reports, three capability tiers of a second model family cover 94.4\% to 78.7\% of the investigators' causal elements, while the share of supported nodes stays near 93\%. Two of these documents are in the corpus, and 104 of their 151 accidents appear in it as database narratives, so the check covers a second document type about largely the same accidents rather than held-out ones.

The extraction is also stable under repetition. Three further extractions of the 2{,}000 evaluation reports at the production temperature of 0.2 give recall between 0.906 and 0.911. The factor sets of a record agree between runs with a mean Jaccard index of 0.80, and the first primary cause node agrees in 85\% of records. A temperature of 0.7 keeps recall at 0.907 and lowers the agreement to 0.72. The temperature therefore governs which secondary factors a chain names and leaves the cause-flagged content in place.

\subsection{Physical Admissibility of the Extracted Chains}\label{sec:violation}

We ran the monitor of Section~\ref{sec:automaton} over the 55{,}940 extracted chains, reading each case's temperature and dewpoint from the coded record. The coverage is small. Of 334{,}611 transitions, 1{,}222 (0.37\%) are carburetor icing onsets with recorded weather, which the guard can check. The fuel guard checks nothing, because fuel on board and flight time are not coded fields, and most transitions concern human, organizational and procedural factors that no physics guard covers.

The carburetor icing guard rejects 2.3\% of the 1{,}222 transitions it checks. The rate rises to 27.7\% with inverted margin signs and to 4.0\% with driver tuples permuted across the guarded work list. It rises to 15.5\% when each chain is paired with the weather of an arbitrary weather-recorded accident. Against that null, the 28 rejections under the true pairing have a binomial tail below $10^{-50}$. The verdicts therefore depend on the pairing of chain and weather, and the 2.3\% the monitor rejects are candidates for extraction error.

\subsection{Calibration of the Physics Layer}\label{sec:calibres}

We executed Eq.~\eqref{eq:calib} for every physics factor of the failure study whose drivers the coded record carries. {} For carburetor icing the link is an isotonic regression of the extracted carburetor icing label on the condensable water $w$ of Eq.~\eqref{eq:icewater}. Every operating point without condensable water therefore takes one value, the floor of the link. The link is fitted on the 39{,}770 NTSB accidents with usable weather, 1{,}205 of them with carburetor icing, and scored out-of-fold over five folds. It reproduces their prevalence of 0.0303 by construction and reaches an out-of-fold AUC of 0.609. The combination of Eq.~\eqref{eq:combine} uses the corpus denominator of 55{,}940 records, of which 1{,}456 carry the factor, a base rate of 0.0260. The link therefore enters it as a relative risk times the corpus base rate, so the worked operating point of Section~\ref{sec:icecf} receives $0.0439\times 0.0260/0.0303=0.0377$. The Brier score falls from 0.50 for the chart probability to 0.0292, against 0.0294 for the base rate. The link is thus barely sharper than the base rate. Its value lies in the ordering and in correcting a level that the chart, a per-exposure zone label, overstates by a factor of 20 to 100. The fuel term is not calibrated, for want of coded fuel and flight time fields, and the failure study labels it as such.

\subsection{How Much Separation the Recorded Weather Can Support}\label{sec:ceiling}

An AUC of 0.61 may be weak, or it may be close to the limit of two weather fields. To decide, we fitted supervised models to the same label from the same inputs and read off their cross-validated performance.

Scored out-of-fold over five folds on the 1{,}205 carburetor icing accidents and 38{,}565 others, a{} logistic regression on temperature and dewpoint reaches 0.642 (0.628, 0.656), and adding the dewpoint spread, relative humidity and quadratic terms leaves it at 0.642. Gradient boosting on the same two inputs reaches 0.623, so no nonlinear structure remains to recover. Because the input is two dimensional, the Bayes-optimal separation can also be estimated directly. It is the AUC of the out-of-fold density ratio of the two classes from kernel density estimates. This gives 0.640 with a 95\% interval of (0.626, 0.655) at a 3~$^\circ$C bandwidth and 0.638 at 2~$^\circ$C. Against that reference the physics model attains 0.612 with a DeLong interval of (0.597, 0.628) while fitting no parameter to the label, which is 80\% of the separation available above chance. The gap of 0.03 to the fitted model is statistically real, with a paired DeLong test giving $z=4.8$, and it is small. The recorded weather cannot distinguish these accidents much more sharply than the physics already does. Most of the residual error belongs to state the database never recorded, chiefly carburetor heat use. One caveat applies to every model. The label is the investigator's attribution, and investigators consult the same published chart, so part of the separation may measure investigator practice. Section~\ref{sec:metar} scores the model against routine weather observations at the same airports, which carry no investigator label.

\subsection{Separation Against Routine Weather}\label{sec:metar}

The evaluations above contrast accidents with accidents. To score the model against weather in which nothing happened, we matched 1{,}033 carburetor icing accidents and 2{,}190 other accidents to the nearest NOAA Integrated Surface Database station, a median of 20 km away. We drew the routine hours of the accident's calendar month there, either all of them or those within three hours of the accident's clock time. Against routine hours the chart probability does not separate the carburetor icing accidents, with an AUC of 0.43 (0.41, 0.45) on any hour and 0.47 (0.46, 0.49) on matched hours, and it separates the other accidents even less, at 0.35 and 0.40 respectively. Accidents of every kind cluster in drier and warmer weather than the month at large, and the chart places 68\% of routine hours in the serious zone. Routine hours are therefore not the exposure of the fleet, since flying concentrates in fair weather.

The ratio of the two accident groups removes that effect. The share of carburetor icing accidents in a chart zone relative to the share of routine hours, divided by the same ratio for the other accidents, is 0.54 (0.40, 0.69) in the nil zone, 0.86 (0.63, 1.11) and 0.97 (0.86, 1.09) in the light and moderate zones and 1.20 (1.15, 1.27) in the serious zone, with intervals from 2{,}000 resamples of the accidents. The shift toward the weather the physics marks is modest, and only the two outer zones differ from one. On the same subset the accident-against-accident AUC is 0.618 (0.598, 0.638). The other accidents stand in for the exposure of the fleet, which is a case-control design with accident controls. A per-flight-hour level would need activity data by weather, which the general aviation activity survey supplies only by year \citep{faa_gasurvey}.

\subsection{Comparison Against Either Layer on Later Years}\label{sec:heldout}

We compare the hybrid with either layer alone on an occurrence task rather than on severity. The physics term carries almost no severity signal. Adding the carburetor icing and wind terms to a logistic model of the extracted factors moves the AUC for a serious or fatal outcome only from 0.818 to 0.820. The carburetor icing term bears on whether a power loss occurs, and the paper makes no claim that the physics layer predicts how the forced landing ends.

The task predicts, from information available before the flight, whether an NTSB accident chain carries the carburetor icing factor, alone or together with engine failure. Records before 2015 fit every fitted arm, and the 8{,}610 records from 2015 onward, with 348 and 335 positives, score them. The arms are (1) the chart physics with no fit, (2) the physics calibrated on the training years, (3) a corpus-only logistic model on month, hour, light, basic weather condition and state, (4) that model with raw temperature and dewpoint, (5) the hybrid, which adds the calibrated physics term to the covariates, and (6) and (7) the hybrid with that term sign inverted or permuted. The hybrid reaches 0.586 on both targets (0.5860 and 0.5856), above the corpus-only arm at 0.553 (paired DeLong $z=4.1$) and the calibrated physics at 0.578. The corrupted hybrids fall back to 0.553, so the gain comes from the physics. The raw feature model reaches 0.597, which a paired test does not separate from the hybrid ($z=1.4$, $p=0.17$), and its lead depends on how recent its training years are. Trained before 2000 it scores 0.555 against 0.554 for the hybrid, while the calibrated physics holds between 0.574 and 0.579 over every window. The time shift costs the raw model 0.057 in AUC and the calibrated physics 0.035. Brier scores sit at the base rate for every arm, because the icing prevalence was 2.8\% in the training years and 4.0\% in the test years and no arm anticipates that drift. On accuracy the hybrid does no better than a fit on the raw weather. Only the calibrated physics is stable across time, since the hybrid inherits the drift of its corpus covariates.

\subsection{Graph-Level Interventions}\label{sec:backdoorres}

Section~\ref{sec:cf} places two kinds of intervention under a single do-operator semantics, and the failure study exercises the physical kind. For the graph-level kind, we take each learned network $G_c$, read the parents of each factor as the adjustment set $Z$, and evaluate Eq.~\eqref{eq:backdoor} with a severe outcome as $Y$. We compare the result with the crude contrast $P(Y\mid X{=}1)-P(Y\mid X{=}0)$ that an analysis without a learned structure would report. The candidate set holds the 39 factors with at least 30 records in the loss of control category and the 27 in the airframe failure category. Of these, 15 and 11 carry a nonempty adjustment set.

Each record enters the category of its first recognized NTSB occurrence code, and neither category matches one CICTT category exactly. Code 250, loss of control in flight, decides 94\% of the loss of control category, which also takes abrupt maneuver, uncontrolled altitude deviation and roll over. Code 130, airframe, component or system failure, decides 97\% of the airframe failure category, which also takes propeller failure and decompression. Its nearest CICTT category is SCF-NP, the failure of a system other than the powerplant, although CICTT counts propeller failure as SCF-PP. Loss of engine power has its own code and falls in neither category, so neither network supplies an adjustment set to the combination of Section~\ref{sec:efcombine}. No category isolates it, since the legacy grouping merges code 350 with three other occurrence codes. Table~\ref{tbl:backdoor} therefore uses the two categories that a single code dominates and reports the graph-level intervention apart from the failure study.

Table~\ref{tbl:backdoor} gives both columns for the factors with a nonempty adjustment set whose adjustment moves the estimate by at least 0.05, with 95\% bootstrap percentiles over 400 resamples for the adjusted effect. The code repository \citep{pang2026repo} lists all 26 factors with a nonempty adjustment set, together with the full networks. In the loss of control category, spatial disorientation appears to raise the probability of a severe outcome by 0.396 before adjustment. After conditioning on low visibility and darkness it raises it by 0.264, which removes a third of the apparent effect. The learned parents are the conditions that produce disorientation in the first place, so the crude figure credits disorientation with harm that the weather and the light already caused. In the airframe failure category, structural failure falls from 0.188 to 0.088 once landing gear anomaly and structural overload are adjusted for. Across the 15 and 11 factors with a nonempty adjustment set, the mean absolute shift is 0.038 and 0.028 respectively, with a maximum of 0.133.

Whether the adjustment moves the estimates toward the truth depends on the causal sufficiency that PC assumes and that Section~\ref{sec:data} argues fails. We therefore ran FCI \citep{spirtes2000causation}, which allows latent confounders and selection, on the same presence data with the same tests and without the ordering priors. Its partial ancestral graphs are dominated by bidirected edges, 47 of 71 adjacencies in the loss of control category and 27 of 45 in the airframe failure category. Of the 29 adjustment relations behind the 15 loss of control factors, FCI admits only 5 as possible parents, and 9 of the 11 airframe failure factors are confounded on its reading. Table~\ref{tbl:backdoor} marks each relation of an adjustment set, in order, as a possible parent (poss), a bidirected edge (bi) or an edge oriented away from the factor (rev), and no relation in its rows is a definite parent. FCI leaves the two parents of spatial disorientation oriented away from it, so their direction rests on the ordering priors and the chain precedence. It marks the relation between landing gear anomaly and structural failure as bidirected, so adjusting on it is not licensed. Recomputed on the relations FCI admits, the structural failure effect is 0.160 rather than 0.088 and the disorientation effect stays at its crude 0.396. The E-values \citep{vanderweele2017sensitivity} on the adjusted effects give the confounder strength that would explain each effect away. They are 2.4 for spatial disorientation and 2.3 for structural failure in the airframe failure category, and 1.5 and 1.8 at the interval limits nearest the null, the values in parentheses in Table~\ref{tbl:backdoor}. Control surface anomaly, whose crude contrast of $-$0.0001 hides an adjusted $-$0.070, also excludes zero, at 2.1 (1.4). The other two intervals, structural failure in the loss of control category and icing encounter, cross zero with E-values of 1.0 at the limit, which no confounding is needed to explain, so the analysis supports only these three rows. The graph-level interventions are estimates on a selected population whose direction rests on the ordering priors, while the physical interventions take their direction from the mechanism.

\begin{tblwide}
\centering
\caption{Graph-level interventions for the occurrence categories of loss of control in flight and airframe failure.}
\label{tbl:backdoor}
\fittab{\begin{tabular}{@{}lrrrlll@{}}
\toprule
Factor & $n$ & Crude & Adjusted (95\% CI) & Adjustment set & FCI & E-value \\
\midrule
\multicolumn{7}{@{}l}{\emph{Loss of control in flight, 3{,}852 records, baseline $P(Y)=0.518$}} \\
\quad Spatial disorientation & 111 & 0.396 & 0.264 (0.058, 0.415) & low visibility, dark night & rev, rev & 2.4 (1.5) \\
\quad Structural failure & 88 & 0.004 & $-$0.088 ($-$0.220, 0.043) & stall, latent defect, overload & bi, bi, bi & 1.7 (1.0) \\
\quad Icing encounter & 45 & 0.165 & 0.096 ($-$0.060, 0.227) & low visibility & bi & 1.7 (1.0) \\
\addlinespace
\multicolumn{7}{@{}l}{\emph{Airframe failure, 1{,}842 records, baseline $P(Y)=0.213$}} \\
\quad Structural failure & 454 & 0.188 & 0.088 (0.045, 0.126) & gear anomaly, overload & bi, poss & 2.3 (1.8) \\
\quad Control surface anomaly & 387 & $-$0.0001 & $-$0.070 ($-$0.116, $-$0.017) & other system failure, gear anomaly & bi, bi & 2.1 (1.4) \\
\bottomrule
\end{tabular}}
\end{tblwide}

\section{Discussion and Limitations}\label{sec:disc}

The main result of this paper is an account of what accident archives can support once a physical model stands behind a factor, and several of the answers are negative. Recorded temperature and dewpoint separate carburetor icing accidents from other accidents only up to a density ratio ceiling of 0.64, and the physics reaches 0.61 without fitting the label. Routine weather hours cannot serve as the control, because flying concentrates in fair weather. On later years the physics adds 0.03 in AUC to the prediction of occurrence and nothing to severity, and a fit on the raw weather matches the hybrid in accuracy. FCI does not license the adjustment that moves the spatial disorientation effect. The physics layer therefore contributes an ordering that holds across time, a stated reference class for every number and interventions that name a physical cause. Its case rests on interpretability, and the paper claims no gain in predictive accuracy. The calibrated numbers refer to a reportable accident, so the reporting bias of the archive is made explicit and confined to the level. The guard verdicts, the rankings across operating points and the ratios between scenarios do not depend on it.

The results share one pattern, in which each layer is informative where the other is silent. The corpus pins down consequences, as the carburetor icing node shows with a sharp edge into engine failure and sparse, weak antecedents. The same division of labor underlies hybrid causal logic and the causal model for air transport safety \citep{groth2010hybrid,ale2009further}. Here the structure comes from report text instead of expert elicitation, so it can be rebuilt as the archive grows.

Four limitations bound these claims. First, the corpus contains only flights that ended in a reportable event, so every calibrated probability and every data-side intervention is conditional on an occurrence, and the selection induces the collider effect of Section~\ref{sec:data}. The carburetor icing evaluation contrasts accidents with accidents under an investigator label that draws on the same published chart, so part of the measured separation may be investigator practice. A per-flight risk needs activity data by weather, with flight hours from the general aviation activity survey as the denominator \citep{faa_gasurvey}.

Second, the fitted leaky noisy-OR carries the causes its set omits in a leak of 8\%, and where an upstream cause acts through a parent already in the set, the per-factor decomposition attributes to that mediator what the upstream cause produced.
A chain-level activation model{} would attribute it upstream, at the cost of the one-level decomposition.

Third, the physical models carry idealizations of their own. The carburetor icing constants come from a published chart, and the level is set by the calibration of Section~\ref{sec:calibres}. The calibrated link is nearly flat across the moderate and serious zones, so it separates little among conducive weathers, and the fuel term is uncalibrated.

Fourth, the same model family performs the extraction, supplies the discovery priors and judges the extraction. The check on 157 full investigation reports varies the extractor, which limits the practical reach of that circularity without removing it. Most of its accidents are also in the corpus, 104 of 151, so it tests a second document type rather than held-out accidents. A blind check by one annotator on 50 reports agrees with the judge on 96.6\% of findings and finds no leniency.

\section{Conclusion}\label{sec:conclusion}

This paper presented a hybrid framework in which large-scale text mining identifies the factors that lead to a failure mode and physical models quantify how likely each failure is. The two parts assemble into a single explainable failure mode risk. One do-operator treatment covers both graph-level backdoor adjustment and mechanistic physical intervention.

We developed engine power loss end to end, with carburetor icing as the worked physics factor, and computed a counterfactual on named accidents through the same models.

The carburetor icing model sits near the separation limit of the recorded weather. On later years the hybrid gains a small, physics-driven margin over the corpus and matches a fit on the raw weather. Only the physics term keeps its ordering across time, since the hybrid inherits the drift of its corpus covariates. Within the limits stated in Section~\ref{sec:disc}, the approach applies to any failure mode whose contributing factors correspond to physical drivers, and it lifts such a mode from one frequency to a risk at the operating point with its reference class stated.

\section*{Acknowledgment}
The authors thank Jiacheng Xie of the Georgia Institute of Technology for carefully reading the manuscript and for comments on aviation safety causal chains. The authors acknowledge the University of Toronto for its support in carrying out this research. Large language models are a component of the method and an object of study. A locally hosted Qwen3.6-35B-A3B \citep{qwen36model} performs the chain extraction, acts as the judge and supplies the ordering priors. It is served by vLLM 0.19.1 with constrained JSON output and no reasoning trace, at temperature 0.2 for extraction and 0 for judging. Claude Opus 5, Claude Sonnet 5 and Claude Haiku 4.5 (identifiers \texttt{claude-opus-5}, \texttt{claude-sonnet-5} and \texttt{claude-haiku-4-5}, accessed on 9 and 10 August 2026) act as alternative extractors in Section~\ref{sec:extractval}. Anthropic's Claude and Claude Code were used to write parts of the analysis code and to draft and revise the text of all sections. The authors verified every numerical result against the data and the archived code, and take full responsibility for the content. The code, prompts, extracted chains, networks, judge outputs and the scripts behind every table and figure are available at \url{https://github.com/YutianPangASU/AviationSafetyReportKnowledgeGraph} \citep{pang2026repo}, and the NTSB, FAA and TSB source records are public.

\bibliographystyle{IEEEtranN-doi}
\bibliography{bstctl,refs}

\end{document}